\documentclass[journal,twoside,web]{ieeecolor}
\usepackage{etoolbox}
\makeatletter
\@ifundefined{color@begingroup}%
{\let\color@begingroup\relax
\let\color@endgroup\relax}{}%
\def\fix@ieeecolor@hbox#1{%
\hbox{\color@begingroup#1\color@endgroup}}
\patchcmd\@makecaption{\hbox}{\fix@ieeecolor@hbox}{}{\FAILED}
\patchcmd\@makecaption{\hbox}{\fix@ieeecolor@hbox}{}{\FAILED}

\usepackage{tmi}
\usepackage{cite}
\usepackage{amsmath,amssymb,amsfonts}
\usepackage{algorithmic}
\usepackage{graphicx}
\usepackage{textcomp}
\usepackage{xcolor}
\usepackage{booktabs}
\usepackage{multirow}
\usepackage{hyperref}
\usepackage{array}
\usepackage{tabularx}
\usepackage{subcaption}
\usepackage[most]{tcolorbox}

\newtcolorbox{problembox}{
    enhanced, breakable,
    colback=gray!7, colframe=gray!55,
    boxrule=0.4pt, arc=2pt,
    left=8pt, right=8pt, top=6pt, bottom=6pt
}

\newcommand{\benchname}{EviPathBench}

\def\BibTeX{{\rm B\kern-.05em{\sc i\kern-.025em b}\kern-.08em
    T\kern-.1667em\lower.7ex\hbox{E}\kern-.125emX}}
\begin{document}

\pagestyle{empty}

\bstctlcite{IEEEtran:BSTcontrol}

\title{\benchname{}: Benchmarking Evidence Acquisition and Reasoning in Vision-Language Models for Whole-Slide Pathology}

\author{
Dankai Liao, 
Tianyi Zhang, 
Yufeng Wu,
Xinyue Zhang,
Qiaochu Xue,
Zeyu Liu, \\
Dachun Zhao, 
Linghan Cai,
and Yueming Jin,~\IEEEmembership{Member, IEEE}
\thanks{This work was supported by the Ministry of Education Tier 1 grant, Singapore (24-1250-P0001), and the Ministry of Education Tier 2 grant, Singapore (T2EP20224-0028). This work was powered by the UnPuzzle \& PuzzleCloud Platform (https://puzzlelogic.com/unpuzzle) and supported by PuzzleLogic Pte Ltd, Singapore.}
\thanks{Dankai Liao and Tianyi Zhang contributed equally to this work. Corresponding Author: Dachun Zhao (e-mail: dachunzhao@126.com), Linghan Cai (e-mail: linghancai@puzzlelogic.com), and Yueming Jin (e-mail: ymjin@nus.edu.sg)}
\thanks{Qiaochu Xue, and Yueming Jin are with the Department of Biomedical Engineering, National University of Singapore, Singapore 117417, Singapore (e-mails: \{dankai.liao, e1352520\}@u.nus.edu, ymjin@nus.edu.sg)}
\thanks{Dankai Liao, Tianyi Zhang and Yueming Jin are with the Department of Electrical and Computer Engineering, National University of Singapore, Singapore 117417 (e-mails: \{e0556735, zhangtianyi\}@u.nus.edu; ymjin@nus.edu.sg).}
\thanks{Yufeng Wu, Xinyue Zhang, Zeyu Liu, and Linghan Cai are with PuzzleLogic Pte Ltd, Singapore 229594 (e-mails: \{yufengwu, xinyuezhang, zeyuliu, linghancai\}@puzzlelogic.com).}
\thanks{Linghan Cai is also with School of Computer Science and Technology, Harbin Institute of Technology, Shenzhen, China (e-mail: linghancai@puzzlelogic.com).}
\thanks{Dachun Zhao is with the Department of Pathology, Peking Union Medical College Hospital, Beijing, China (e-mail: dachunzhao@126.com).}}

\maketitle
\thispagestyle{empty}

\begin{abstract}
Whole-slide image (WSI) diagnosis requires identifying diagnostically relevant regions, examining them across magnifications, and integrating multi-scale evidence. However, most pathology benchmarks evaluate models on pre-cropped patches or pre-extracted slide features, leaving their ability to acquire evidence from gigapixel WSIs largely untested.
We introduce \benchname{}, a benchmark for evaluating evidence acquisition and reasoning in vision-language models (VLMs) for whole-slide pathology. It evaluates four capabilities: image-to-text matching for evidence interpretation, text-to-image retrieval for evidence verification, diagnostic-region localization for evidence acquisition, and multi-scale reasoning for evidence integration. The benchmark is organized as a diagnostic tree linking nested regions across magnifications with scale-specific findings and path-level diagnoses. It contains 1{,}822 TCGA WSIs and 17{,}135 diagnostic paths annotated by ten board-certified pathologists. A private cohort of 190 breast cancer WSIs with detailed annotations further evaluates autonomous whole-slide exploration.
We evaluate 19 VLMs spanning general-purpose, medical, and pathology-specialized families, plus one text-only reference model. Leading open-weight models achieve over 93\% accuracy in multi-scale reasoning and over 50\% in both cross-modal matching tasks. In contrast, diagnostic-region localization remains challenging: the best text-guided mean intersection-over-union is below 0.09, underperforming a center-based heuristic. During autonomous exploration, the unconditional hit rate drops from 0.522 at low magnification to 0.185 at intermediate magnification and 0.020 at high magnification. These results reveal a pronounced gap between reasoning over curated evidence and acquiring it from WSIs. \benchname{} provides a unified framework for measuring and improving both capabilities.
The benchmark is accessible at \mbox{\href{https://github.com/dankail/EviPathBench}{GitHub}}.
\end{abstract}
\begin{IEEEkeywords}
Whole-slide image, computational pathology, vision-language model, evidence acquisition, multi-scale reasoning
\end{IEEEkeywords}

\section{Introduction}
\label{sec:introduction}
\IEEEPARstart{P}{athological} diagnosis from whole-slide images (WSIs) is inherently an evidence-seeking process. A pathologist first surveys the slide at low magnification to identify suspicious tissue, then examines selected regions at progressively higher magnifications. Observations of tissue architecture, growth patterns, and cellular morphology are subsequently integrated into a diagnostic conclusion~\cite{chakrabortyDecodingVisualAttention2024}. Reliable WSI analysis therefore requires more than interpreting a given image region. It also requires deciding where to look, which regions to examine further, and how to combine evidence acquired across locations and magnifications.

The need to evaluate this complete workflow has become increasingly pressing as vision-language models (VLMs) are applied to computational pathology. General-purpose models can interpret histopathology patches, while pathology-specialised VLMs support morphology recognition, visual question answering, and diagnostic dialogue~\cite{luMultimodalGenerativeAI2024}. More recently, VLM-based WSI systems have incorporated tool use, hierarchical zooming, and language-guided planning to support iterative examination of gigapixel slides~\cite{chenPathAgentInterpretableAnalysis2025}. Together, these developments mark a transition from passive interpretation of selected images toward active examination of gigapixel slides. Yet, the evaluation of these systems has not progressed at the same pace.

Despite this shift toward active WSI examination, most existing pathology benchmarks assess models only after the relevant visual evidence has already been selected. Patch-level benchmarks provide a cropped field of view~\cite{sunPathMMUMassiveMultimodal2024}, while slide-level benchmarks commonly rely on pre-extracted features, aggregated representations, or manually selected regions~\cite{Liang_2025_ICCV}. These settings measure whether a model can interpret supplied evidence, but not whether it can acquire that evidence from a WSI. This distinction is critical because a model may reason accurately over a diagnostic region while remaining unable to locate that region independently. Consequently, strong performance on conventional visual question answering or slide-level reasoning does not establish that a model can perform end-to-end diagnostic exploration.

\begin{figure}[t]
\centering
\makebox[\columnwidth][c]{%
\includegraphics[width=1.08\columnwidth]{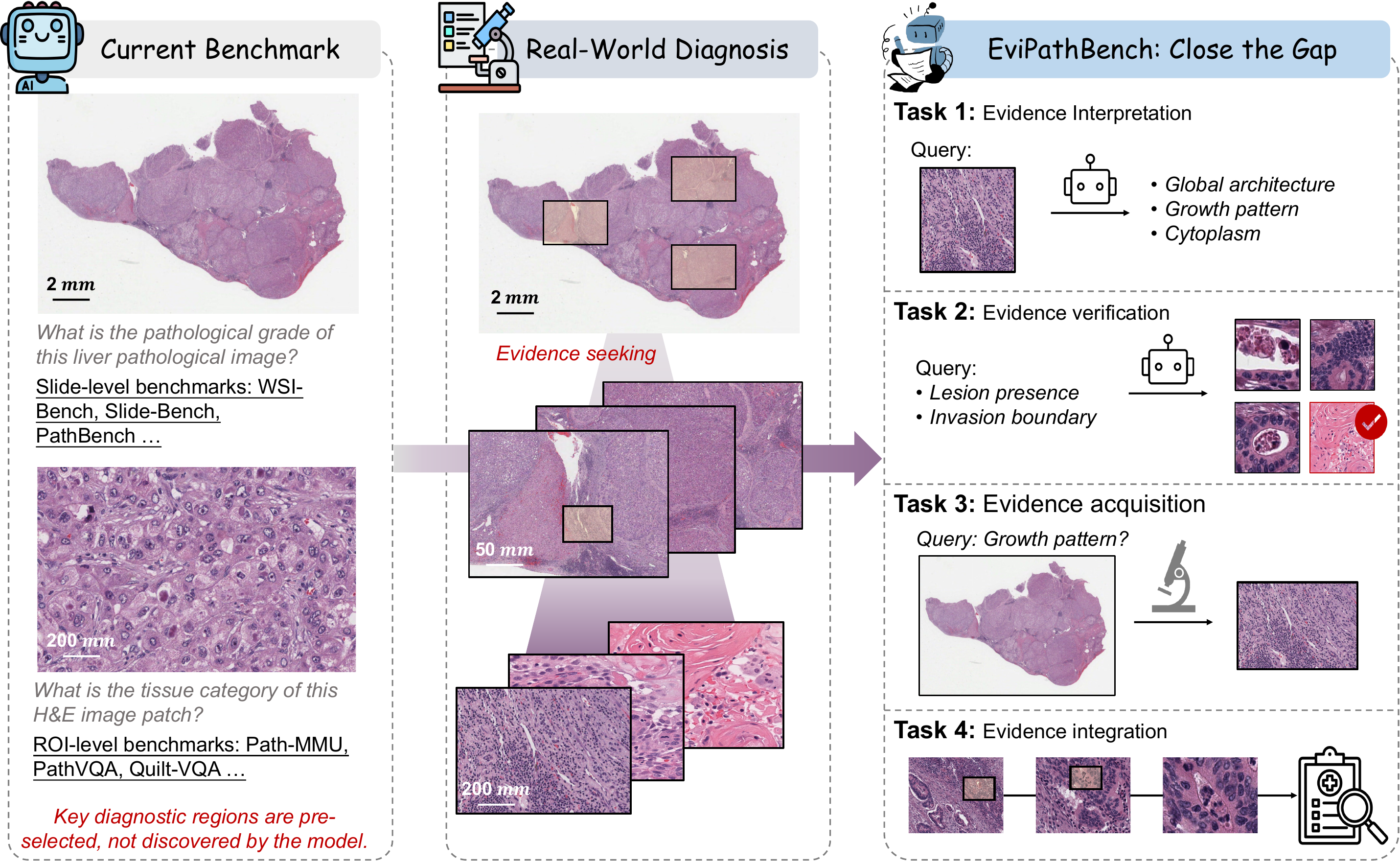}}
\caption{Motivation for \benchname{}. Current pathology benchmarks predominantly evaluate interpretation of preselected evidence, whereas real-world WSI diagnosis requires active evidence seeking across locations and magnifications. \benchname{} is designed to bridge this gap.}
\label{fig:motivation}
\vspace{-4pt}
\end{figure}

To make this missing capability explicit and measurable, we formulate WSI examination as a hierarchical, multi-magnification diagnostic tree. At low magnification, the model should survey the global tissue context and localize suspicious regions; at intermediate magnification, it should assess tissue architecture and growth patterns; and at high magnification, it should inspect cellular and nuclear morphology. Each region-selection decision constrains which areas are examined at subsequent stages. In this formulation, the root corresponds to the complete slide, each node represents a region viewed at a particular magnification, and each branch encodes a possible diagnostic traversal. Diagnostic performance therefore depends on two coupled capabilities: selecting an informative path through the tree and reasoning over the evidence accumulated along that path. Existing benchmarks primarily assess the latter, while largely overlooking the former---the ability to navigate the slide and acquire diagnostically informative evidence.

Building on this distinction between evidence acquisition and reasoning, we introduce \benchname{}, a benchmark designed to evaluate both processes within a unified diagnostic framework. As illustrated in Figs. 1 and 2, \benchname{} represents four stages of WSI diagnosis as four corresponding benchmark tasks: evidence interpretation through image-to-text matching, evidence verification through text-to-image retrieval, evidence acquisition through diagnostic region localization, and evidence integration through multi-scale diagnostic reasoning. Image-to-text matching assesses whether a model can interpret the morphology within a diagnostic region, whereas text-to-image retrieval tests whether it can verify a textual hypothesis against candidate image evidence. Diagnostic region localization evaluates evidence acquisition in two complementary settings: text-guided localization and autonomous whole-slide exploration. Finally, multi-scale diagnostic reasoning assesses whether findings collected across magnifications can be integrated into a coherent diagnosis. Evaluating these tasks jointly reveals where a model fails within the diagnostic workflow, rather than reducing its performance to a single aggregate score.

To instantiate this framework at scale, \benchname{} draws on 2,012 WSIs from The Cancer Genome Atlas (TCGA)~\cite{thecancergenomeatlasresearchnetworkCancerGenomeAtlas2013} and an in-house breast cohort, spanning 16 organ types, together with 17,135 diagnostic paths selected by ten board-certified pathologists. Each path contains nested bounding boxes, findings at three magnifications, and a path-level diagnostic conclusion. Using this benchmark, we evaluate 20 model configurations---19 VLMs spanning general-purpose, general-medical, and pathology-specialised families, plus one text-only reference model---across the four stages. This evaluation reveals a clear capability asymmetry. Leading models achieve over 93\% accuracy when integrating pre-specified multi-scale findings, yet all models perform poorly when required to locate diagnostic evidence. Even the strongest closed-source models obtain a mean intersection-over-union below 0.09 in text-guided localization, while autonomous exploration suffers substantial coverage loss across successive magnifications. These findings identify evidence acquisition, rather than reasoning over preselected evidence, as the principal bottleneck in current VLM-based WSI systems. The main contributions of this work are threefold.
\begin{itemize}
    \item We formulate VLM evaluation for whole-slide pathology on a multi-scale diagnostic tree, explicitly separating evidence acquisition from reasoning over evidence.
    \item We present \benchname{}, which instantiates interpretation, verification, acquisition, and integration as four complementary evaluation tasks using pathologist-authored annotations.
    \item Extensive evaluation and analyses on \benchname{} show that current pathology vision-language models can interpret and integrate supplied evidence but remain unreliable at spatially grounded, multi-step WSI navigation.
\end{itemize}

\begin{figure*}[t]
\centering
\includegraphics[width=\textwidth]{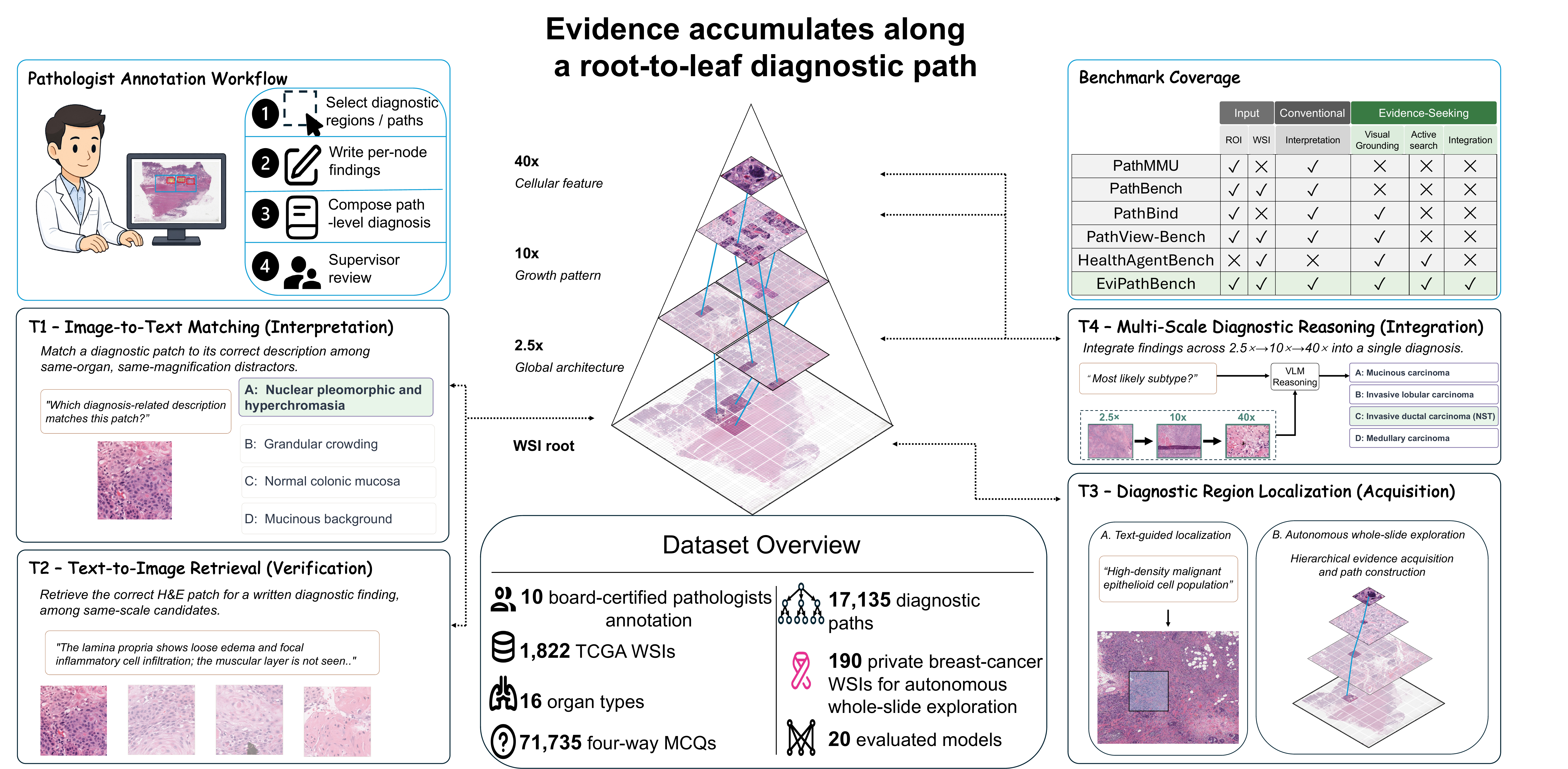}
\caption{Diagnostic-tree formulation of evidence seeking. A WSI is the root; at each magnification, selected regions become child nodes for further zoom-in while the remainder is pruned. Paths accumulate evidence through interpretation, verification, acquisition, and integration, instantiated by the four \benchname{} tasks.}
\label{fig:diagnostic_tree}
\end{figure*}

\section{Related Work}
\label{sec:related_work}
\subsection{Vision-Language Models for Pathology}
General-purpose VLMs have substantially improved multimodal perception, instruction following, and structured reasoning~\cite{openai_update_2025,googledeepmindGemini3Flash2025,xaiGrok4Model2025}. Open-weight model families have further extended these capabilities across a wide range of parameter scales and computational budgets~\cite{chenExpandingPerformanceBoundaries2025,qwenteamQwen35NativeMultimodal2026,baiQwenVLVersatileVisionLanguage2023,teamKimiK25Visual2026,grattafioriLlama3Herd2024,abdinPhi3TechnicalReport2024,Liu_2024_CVPR,li2024llavanextstrong}. General-medical models, including LLaVA-Med~\cite{NEURIPS2023_5abcdf8e}, MedGemma~\cite{sellergrenMedGemmaTechnicalReport2026}, and Lingshu~\cite{lasateamLingshuGeneralistFoundation2025}, introduce medical-domain supervision while
retaining the broad instruction-following capabilities of general-purpose models. However, these advances are primarily demonstrated on images or image collections that can be directly presented to the model. Their applicability to WSIs remains constrained by the gigapixel scale and multi-resolution structure of slides.

Medical VLMs applied to pathology can be broadly divided according to the spatial scale at which evidence is processed. Patch-level models learn representations or generate responses from individual fields of view. Representative systems include PathChat~\cite{luMultimodalGenerativeAI2024}, PathAsst~\cite{sunPathAsstGenerativeFoundation2024}, Quilt-LLaVA~\cite{Seyfioglu_2024_CVPR}, LLaVA-Med~\cite{NEURIPS2023_5abcdf8e}, and Patho-R1~\cite{zhangPathoR1MultimodalReinforcement2026}. These models have demonstrated promising capabilities in morphology recognition, visual question answering, and diagnostic dialogue. In contrast, slide-level models such as SlideChat~\cite{Chen_2025_CVPR}, WSI-LLaVA~\cite{Liang_2025_ICCV}, and PathReasoner~\cite{jiangPathReasonerR1InstillingStructured2026} aggregate information from multiple regions to support WSI-level question answering or diagnostic reasoning. Patch-level models therefore emphasize local evidence interpretation, whereas slide-level models emphasize the integration of distributed evidence. Together, they cover the interpretation and integration endpoints of WSI analysis, but do not by themselves characterize the intervening process of region selection.

Several recent VLM-based WSI systems further support iterative coarse-to-fine inspection through region selection, hierarchical zooming, or tool-mediated querying. CPathAgent~\cite{sunCPathAgentAgentbasedFoundation2025} and PathAgent~\cite{chenPathAgentInterpretableAnalysis2025} model coarse-to-fine diagnostic trajectories, while MMNavAgent~\cite{xuMMNavAgentMultiMagnificationWSI2026} and PathFound~\cite{huaPathFoundAgenticMultimodal2026} emphasize multi-magnification navigation and active evidence acquisition. Complementing these systems, Pathology-CoT derives navigation and behavior-guided reasoning supervision from expert WSI viewing trajectories~\cite{wangPathologyCoTLearningVisual2026}, whereas PathScale-R1 targets shortcut-resistant reasoning across magnifications~\cite{phanPathScaleR1CrossscaleReasoning2026}. These studies establish the feasibility of iterative WSI examination, but their heterogeneous tasks, cohorts, and endpoints make it difficult to separate acquisition from downstream reasoning or compare VLMs under a common protocol.

\subsection{Benchmarks for Pathology Image Understanding}
Patch-level benchmarks such as PathVQA~\cite{hePathVQA30000Questions2020}, Quilt-VQA~\cite{Seyfioglu_2024_CVPR}, PathMMU~\cite{sunPathMMUMassiveMultimodal2024}, and OmniPathoVQA~\cite{chenOmniPathoVQABenchmarkingPathology2026} evaluate preselected regions. WSI-VQA~\cite{chenWSIVQAInterpretingWhole2024}, SlideBench~\cite{Chen_2025_CVPR}, WSI-Bench~\cite{Liang_2025_ICCV}, and PathBench~\cite{sunPathBenchAdvancingBenchmark2025} evaluate aggregated WSI representations with region selection fixed before inference. Recent work instead asks whether predictions are supported by visual evidence. PathBind~\cite{zhangPathologyVisionLanguageModels2026} filters text-solvable VQA samples and evaluates visual dependence and region-level grounding. PathView-Bench~\cite{chenPathViewBenchCanMultimodal2026} extends this direction through vision-anchored localization, recognition, quantification, spatial reasoning, and evidence-sufficiency tasks across fixed region- and slide-level views. Together, they assess visual grounding and multiscale understanding. However, neither captures how a VLM searches a WSI for diagnostic evidence. HealthAgentBench~\cite{liuHealthAgentBenchUnifiedBenchmark2026} moves toward interactive WSI exploration, but pathology is represented by a single tumor-localization task. No existing benchmark spans interpretation, verification, acquisition, and integration within a common framework. \benchname{} provides this stage-wise evaluation using pathologist-annotated diagnostic trees.

\section{\benchname{} Construction}
\label{sec:methods}

In this section, we first describe the diagnostic-tree representation, data sources, annotation workflow, and then introduce the evaluation protocols used to construct \benchname{}.

\begin{figure*}[t]
\centering
\includegraphics[width=\textwidth]{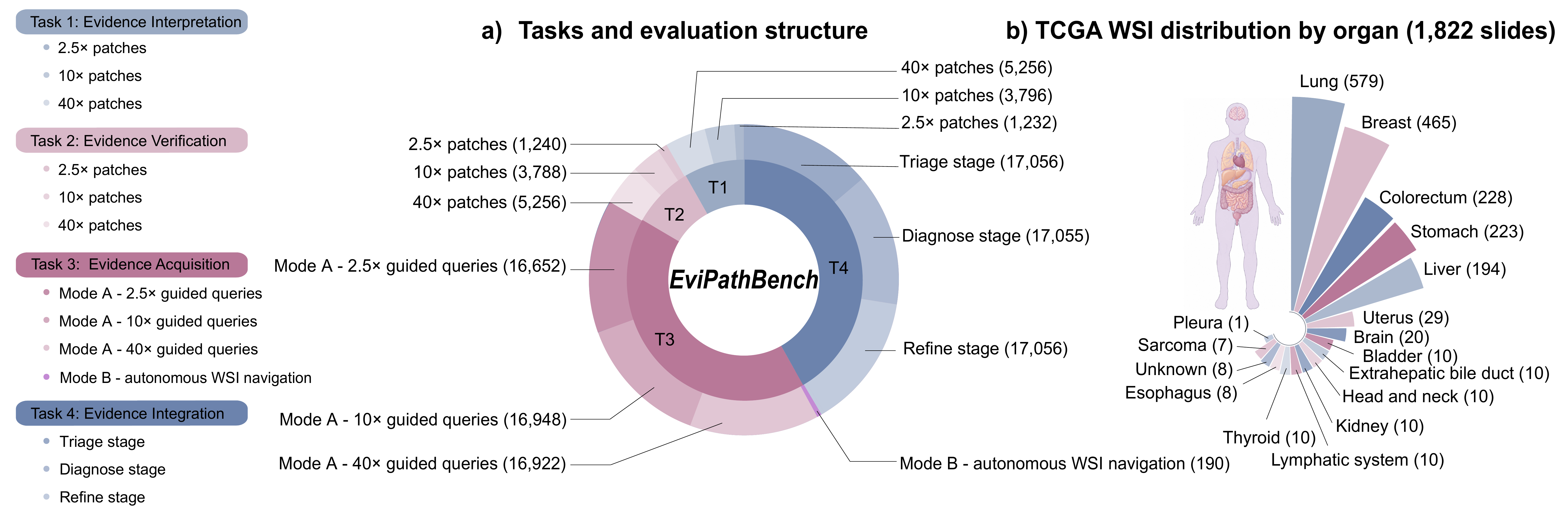}
\caption{Overview of \benchname{}. (a) Four tasks and their evaluation axes; numbers denote task-specific evaluation budgets. (b) TCGA WSI distribution by organ (top 10 plus Other). The separate private breast cohort used for Mode~B is not shown.}
\label{fig:benchmark_overview}
\end{figure*}

\subsection{Diagnostic Tree and Acquisition--Reasoning Decomposition}
\label{sec:tree}

A WSI $s$ induces a diagnostic tree $\mathcal{T}_s = (\mathcal{V}_s, \mathcal{E}_s)$ over its image pyramid. Each node $v = (s, \ell, b)$ pairs a magnification $\ell \in \mathcal{L} = \{2.5\times, 10\times, 40\times\}$ with a box $b \subseteq \Omega_s$ on the full-slide canvas; the root $r_s = (s, \perp, \Omega_s)$ is the thumbnail. Every non-root node $v$ has a unique annotated parent $\operatorname{pa}_s(v)$ at the preceding view, with $\mathrm{next}$ defined by $\perp \mapsto 2.5\times \mapsto 10\times \mapsto 40\times$ and $c(b(v))\in b(\operatorname{pa}_s(v))$ for box centre $c(b)$. Thus,
\begin{equation}
    \mathcal{E}_s=\bigl\{(\operatorname{pa}_s(v),v):v\in\mathcal{V}_s\setminus\{r_s\}\bigr\},
    \label{eq:tree_edges}
\end{equation}
which remains a tree even when overlapping boxes admit more than one geometric container. A root-to-leaf path $\pi = (r_s, v_1, v_2, v_3)$ with $(\ell(v_1),\ell(v_2),\ell(v_3))=(2.5\times,10\times,40\times)$ represents one diagnostic search route; each node carries a pathologist-authored finding $d(v)$ and each path carries a pathologist-composed summary $D(\pi)$.

Existing patch- and slide-level VQA benchmarks measure reasoning given evidence that has already been supplied. To formalise the additional search component, let $\Pi$ denote a single path sampled from a model's diagnostic policy and let $Y$ denote its diagnostic output. Marginalising over the latent path gives
\begin{equation}
    p_\theta(y \mid s) = \sum_{\pi \in \mathcal{P}(\mathcal{T}_s)}
    p_\theta(y \mid \Pi=\pi,s)\,p_\theta(\Pi=\pi \mid s).
    \label{eq:decomposition}
\end{equation}
The two terms represent reasoning over evidence and evidence acquisition, respectively; a set-valued $\Pi$ gives the corresponding multi-branch formulation. Conventional VQA fixes $\Pi$ implicitly, whereas \benchname{} evaluates both components.

\subsection{Data Sources}
\label{sec:data_sources}

\benchname{} draws on two complementary sources. The first is a TCGA-based diagnostic tree dataset of 1{,}822 WSIs supporting tasks 1, 2, 3A, and 4. Of these, 1{,}814 carry one of 16 standardised organ labels (Lung, Breast, Colorectum, Stomach, Liver dominate); the remaining 8 are labelled ``Unknown''. The per-organ distribution and task structure are shown in Fig.~\ref{fig:benchmark_overview}. The second is a private 190-WSI breast-cancer cohort for Mode~B; each slide has a clinically annotated tumor polygon for Nottingham grading but no multi-scale path structure, precluding Mode~A evaluation.

\subsection{Pathologist Annotation Workflow}
\label{sec:annotation}

All per-node bounding boxes, per-magnification findings, and path-level diagnoses in the TCGA dataset are pathologist-authored. Ten board-certified pathologists used an internal tool to select diagnostic regions and record the unique parent relation in Eq.~\eqref{eq:tree_edges}. To qualitatively assess cross-annotator consistency, all ten pathologists independently annotated a shared calibration subset of 50 WSIs. Supervising review found that most annotators identified the principal diagnostically relevant paths across this subset. Each remaining slide was assigned to one primary annotator, who selected the critical root-to-leaf paths $\pi$, wrote a finding $d(v)$ for every node along each path, and composed a path-level summary $D(\pi)$. Across both groups, a supervising pathologist reviewed the spatial selections, findings, and summaries; flagged content was returned to the annotator and the slide re-entered review until it passed. Only passed slides are released. Only passed slides are released. The workflow yielded 17{,}135 paths and 51,363 per-magnification node descriptions over the 1{,}822 WSIs.

\subsection{Evaluation}
\label{sec:task_construction}

\benchname{} formulates the WSI diagnostic workflow as four research questions about VLMs. Each task directly probes a component defined in \S\ref{sec:tree}: Tasks~1, 2, and 4 hold annotated evidence fixed and evaluate reasoning over evidence, whereas Task~3 evaluates evidence-acquisition behavior. In Tasks~1, 2, and 4, all four options are presented jointly and the parsed choice is scored by exact-match accuracy.

\subsubsection{Evidence Interpretation}

\begin{problembox}
\noindent\textbf{Q.} Can a vision-language model read the pathological content of a single node of the diagnostic tree?
\end{problembox}

As a node-level probe of the reasoning term in Eq.~\eqref{eq:decomposition}, \emph{image-to-text matching} presents a node image with four candidate descriptions and asks the model to select the single pathologist-authored finding. All four candidates in a question come from the same organ--magnification cell of the diagnostic tree so that neither tissue type nor magnification leaks the answer. Distractors are mined by encoding each finding in the cell using the Sentence-Transformers model \texttt{all-MiniLM-L6-v2}~\cite{sentence_transformers_all_minilm_l6_v2, reimersSentenceBERTSentenceEmbeddings2019} and greedily selecting the next candidate that minimises the maximum cosine similarity against already-chosen items, rejecting any set whose internal pairwise similarity exceeds $0.6$. Each four-element set yields four questions by rotating the image slot; letter positions are shuffled under a fixed seed; magnification prefixes (``$40\times$:'') are stripped to remove a trivial leakage cue. The released task contains $10{,}284$ four-option questions across the three magnifications.

\subsubsection{Evidence Verification}

\begin{problembox}
\noindent\textbf{Q.} Can a vision-language model verify whether a textual finding matches the morphology actually present in a tissue region?
\end{problembox}

As the complementary node-level probe of the same reasoning term, \emph{text-to-image retrieval} presents a pathologist-authored finding with four candidate images and asks the model to select the matching region. The four-element pool is mined with the same group-and-greedy procedure as T1 so the distractor difficulty is comparable across the two modality directions and any T1$\leftrightarrow$T2 gap reflects a directional alignment property of the model. The textual stem is LM-rewritten into ``Which of the following images shows\ldots'' phrasing for natural medical wording. The released task contains $10{,}284$ four-option questions.

\begin{table*}[t]
\centering
\caption{Per-model performance across the four \benchname{} tasks. T1/T2/T4 report accuracy (\%); T3 reports mean IoU (Mode~A) or conditional and unconditional hit rate (Mode~B).}
\label{tab:results}
\renewcommand{\arraystretch}{0.98}
\fontsize{6.0pt}{6.0pt}\selectfont
\setlength{\tabcolsep}{0.2pt}
\begin{tabular*}{\textwidth}{@{\extracolsep{\fill}} l c c c c c c}
\toprule
\textbf{Model (scale)} & \textbf{T1: image-to-text} & \textbf{T2: text-to-image} & \textbf{T3-A} & \textbf{T3-B cond.} & \textbf{T3-B uncond.} & \textbf{T4: diagnostic reasoning} \\
& Acc. (\%) & Acc. (\%) & mean IoU & hit (\%) & hit (\%) & Acc. (\%) \\
\cmidrule(lr){2-2}\cmidrule(lr){3-3}\cmidrule(lr){4-4}\cmidrule(lr){5-5}\cmidrule(lr){6-6}\cmidrule(lr){7-7}
& 2.5/10/40$\,|\,$All & 2.5/10/40$\,|\,$All & 2.5/10/40 & 2.5/10/40 & 2.5/10/40 & Tri/Diag/Ref$\,|\,$All \\
\midrule
\multicolumn{7}{@{}l}{\textit{Trivial baselines}} \\
Random Choice & 25.00/25.00/25.00$\,|\,$25.00 & 25.00/25.00/25.00$\,|\,$25.00 & ---/---/--- & ---/---/--- & ---/---/--- & 25.00/25.00/25.00$\,|\,$25.00 \\
Frequent Choice & 27.92/26.03/26.22$\,|\,$26.35 & 27.42/26.16/25.36$\,|\,$25.78 & ---/---/--- & ---/---/--- & ---/---/--- & 25.00/25.00/25.00$\,|\,$25.00 \\
\midrule
\multicolumn{7}{@{}l}{\textit{Human reference}} \\
Expert Performance & 92.5/93.8/94.5$\,|\,$93.6 & 91.0/92.4/93.6$\,|\,$92.3 & ---/---/--- & ---/---/--- & ---/---/--- & 98.0/95.5/96.2$\,|\,$96.6 \\
\midrule
\multicolumn{7}{@{}l}{\textit{Non-VLM navigation reference baselines}} \\
Random sampling & ---/---/---$\,|\,$--- & ---/---/---$\,|\,$--- & 0.038/0.073/0.031 & 33.47/16.62/7.40 & 33.47/9.16/0.87 & ---/---/---$\,|\,$--- \\
Grid centroid & ---/---/---$\,|\,$--- & ---/---/---$\,|\,$--- & 0.049/0.146/0.163 & ---/---/--- & ---/---/--- & ---/---/---$\,|\,$--- \\
Tissue density & ---/---/---$\,|\,$--- & ---/---/---$\,|\,$--- & 0.072/0.033/0.021 & 28.60/17.87/6.42 & 28.60/9.68/1.04 & ---/---/---$\,|\,$--- \\
Parent-box center & ---/---/---$\,|\,$--- & ---/---/---$\,|\,$--- & 0.065/0.247/0.283 & ---/---/--- & ---/---/--- & ---/---/---$\,|\,$--- \\
CONCH retrieval & ---/---/---$\,|\,$--- & ---/---/---$\,|\,$--- & 0.038/0.105/0.060 & 30.08/18.30/9.04 & 30.08/7.32/0.67 & ---/---/---$\,|\,$--- \\
\midrule
\multicolumn{7}{@{}l}{\textit{General-purpose models (closed-source)}} \\
GPT-5.2 (2--5\,T$^\ddagger$) & 41.56/50.08/56.28$\,|\,$52.23 & 50.16/58.05/65.60$\,|\,$\underline{60.96} & \textbf{0.086}/0.059/0.066 & 46.81/22.50/7.94 & 46.81/16.42/\underline{1.83} & 96.99/79.37/95.43$\,|\,$89.20 \\
Gemini-3-Flash ($\sim$1.2\,T$^\ddagger$) & 57.31/63.94/64.59$\,|\,$\textbf{63.48} & 62.82/66.53/69.77$\,|\,$\textbf{67.74} & 0.041/\textbf{0.089}/0.068 & \textbf{52.17}/\textbf{24.08}/8.43 & \textbf{52.17}/\textbf{18.48}/\textbf{2.02} & 97.64/82.62/95.95$\,|\,$91.18 \\
\midrule
\multicolumn{7}{@{}l}{\textit{General-purpose models (open-weight)}} \\
Kimi-K2.5 (1\,T / 32\,B-A) & 51.14/56.74/60.67$\,|\,$\underline{58.08} & ---/---/---$\,|\,$--- & ---/---/--- & ---/---/--- & ---/---/--- & ---/---/---$\,|\,$--- \\
DeepSeek-V4-Flash$^\dagger$ (284\,B / 13\,B-A) & ---/---/---$\,|\,$--- & ---/---/---$\,|\,$--- & ---/---/--- & ---/---/--- & ---/---/--- & 96.34/83.64/93.76$\,|\,$90.49 \\
Qwen-3.5-Flash (35\,B / 3\,B-A) & 45.05/52.77/57.69$\,|\,$54.36 & 50.00/57.26/64.44$\,|\,$60.05 & 0.051/0.079/\textbf{0.083} & 42.32/20.33/7.77 & 42.32/14.57/1.55 & 97.10/83.31/95.43$\,|\,$90.98 \\
InternVL2.5-26B (26\,B) & 32.87/36.67/40.62$\,|\,$38.23 & 31.21/33.63/38.64$\,|\,$35.90 & ---/---/--- & 46.92/23.32/\underline{8.52} & 46.92/15.28/1.58 & 97.78/90.01/93.73$\,|\,$93.16 \\
LLaMA-3.2-11B-Vision (11\,B) & 26.30/26.82/27.80$\,|\,$27.26 & ---/---/---$\,|\,$--- & ---/---/--- & ---/---/--- & ---/---/--- & 96.92/80.41/91.87$\,|\,$88.86 \\
InternVL2.5-8B (8\,B) & 33.69/39.01/41.97$\,|\,$39.89 & 29.19/29.91/33.49$\,|\,$31.65 & ---/---/--- & 34.07/16.98/6.84 & 34.07/10.64/1.06 & 97.63/90.89/93.43$\,|\,$\textbf{93.44} \\
LLaMA3-LLaVA-Next-8B (8\,B) & 24.19/26.55/26.20$\,|\,$26.09 & 27.10/26.48/24.03$\,|\,$25.30 & ---/---/--- & 7.03/6.41/4.84 & 7.03/0.62/0.04 & 97.34/76.43/89.46$\,|\,$85.20 \\
LLaVA-1.5-7B (7\,B) & 28.00/26.26/26.05$\,|\,$26.36 & 26.85/26.06/25.06$\,|\,$25.64 & ---/---/--- & 37.30/18.09/7.82 & 37.30/10.77/1.03 & 42.21/30.92/47.28$\,|\,$40.59 \\
Qwen2.5-VL-7B (7\,B) & 28.57/32.35/34.93$\,|\,$33.22 & 31.69/34.77/39.86$\,|\,$37.00 & ---/---/--- & 25.41/17.70/7.94 & 25.41/7.56/0.81 & 95.25/85.29/86.12$\,|\,$88.26 \\
Phi-3-Vision-4.2B (4.2\,B) & 24.43/25.55/25.61$\,|\,$25.45 & ---/---/---$\,|\,$--- & ---/---/--- & 11.11/9.01/7.74 & 11.11/1.79/0.16 & 93.72/90.45/76.52$\,|\,$86.24 \\
\midrule
\multicolumn{7}{@{}l}{\textit{Pathology-specialised VLMs}} \\
Patho-R1-7B (7\,B) & 20.29/19.10/20.97$\,|\,$20.20 & 21.61/22.10/21.80$\,|\,$21.89 & 0.000/0.000/0.000 & 12.62/8.62/6.11 & 12.62/2.23/0.22 & 81.82/60.85/73.23$\,|\,$70.08 \\
Quilt-LLaVA-7B (7\,B) & 6.25/6.06/7.91$\,|\,$7.03 & 6.53/7.55/6.22$\,|\,$6.75 & ---/---/--- & 37.83/18.15/8.36 & 37.83/10.38/1.01 & 69.02/35.08/63.88$\,|\,$56.02 \\
\midrule
\multicolumn{7}{@{}l}{\textit{General-medical VLMs}} \\
Lingshu-32B (32\,B) & 40.67/46.02/49.89$\,|\,$47.36 & 36.94/42.05/47.51$\,|\,$44.22 & ---/---/--- & ---/---/--- & ---/---/--- & 97.55/83.34/94.33$\,|\,$90.65 \\
MedGemma-27B (27\,B) & 27.76/27.08/28.54$\,|\,$27.91 & 22.42/22.62/24.89$\,|\,$23.76 & ---/---/--- & \underline{49.16}/\underline{24.05}/\textbf{8.70} & \underline{49.16}/16.52/1.80 & 97.34/85.88/93.98$\,|\,$91.83 \\
Lingshu-7B (7\,B) & 44.24/51.32/56.72$\,|\,$53.23 & 47.10/51.45/58.43$\,|\,$54.49 & ---/---/--- & 30.47/17.06/6.99 & 30.47/10.01/1.04 & 97.64/90.62/93.36$\,|\,$\underline{93.41} \\
LLaVA-Med-1.5-7B (7\,B) & 20.37/20.44/20.55$\,|\,$20.49 & 18.79/18.66/18.61$\,|\,$18.65 & ---/---/--- & ---/---/--- & ---/---/--- & 72.16/57.27/35.67$\,|\,$53.97 \\
MedGemma-1.5-4B (4\,B) & 30.52/34.51/39.08$\,|\,$36.37 & 24.44/30.62/32.61$\,|\,$30.89 & ---/---/--- & 46.92/22.43/8.27 & 46.92/15.11/1.58 & 89.61/73.27/76.09$\,|\,$78.13 \\
MedGemma-4B (4\,B) & 29.14/31.09/32.61$\,|\,$31.63 & 29.03/32.89/34.02$\,|\,$33.00 & ---/---/--- & 48.90/23.11/8.50 & 48.90/\underline{17.01}/1.80 & 96.62/83.63/89.97$\,|\,$89.06 \\
\bottomrule
\end{tabular*}
\vspace{2pt}
{\tiny
\textbf{Bold} and \underline{underlining} mark the best and second-best model per eligible column; ``---'' denotes not evaluated. $^\dagger$ Text-only LLM. $^\ddagger$ Analyst estimate. T3A uses a fixed 50-slide subset with three queries per slide (150 queries per model); T3B uses the full 190-slide cohort.
}
\end{table*}

\subsubsection{Evidence Acquisition}
\label{sec:task3}

\begin{problembox}
\noindent\textbf{Q.} Can a vision-language model acquire diagnostically relevant evidence by navigating a WSI across scales under different levels of guidance?
\end{problembox}

This task directly evaluates evidence acquisition in Eq.~\eqref{eq:decomposition}. Accordingly, we design two complementary modes that differ in the guidance available to the model. Mode~A supplies an explicit textual target and asks a tool-using VLM to localize the corresponding evidence, whereas Mode~B provides no target-specific guidance and uses the VLM to rank which regions are worth examining within a fixed hierarchy.

\paragraph*{Mode A: text-guided localization.}
Given a slide and a target description, the tool-using VLM returns a predicted box at the requested magnification. We report mean intersection-over-union (mIoU) with the pathologist-annotated box and the fraction of queries with IoU at least $0.3$. For cases with multiple valid annotated boxes sharing the same slide, magnification, query, and parent boundary, IoU is computed against each valid box and the maximum is used for scoring. A LangGraph state machine exposes three tools---\texttt{get\_image\_info()}, \texttt{extract\_roi(x,y,w,h,reason)} in level-0 pixels, and \texttt{finish\_and\_report(conclusion, bbox, confidence)}---under an exploration budget of $(10, 7, 5)$ ROIs at $(2.5\times, 10\times, 40\times)$. At 10$\times$ and 40$\times$, the search space is constrained to the annotated parent box to follow the hierarchy in Eq.~\eqref{eq:tree_edges}. The predicted box is constrained to the ground-truth dimensions so IoU isolates position rather than extent. The evaluation uses a 50-slide stratified subset with three queries per slide, identical across models, together with five non-VLM baselines and an exploration-budget sweep (Table~\ref{tab:ablation_t3a}).

\paragraph*{Mode B: whole-slide diagnostic exploration.}
The VLM scores all 2.5$\times$ tiles with the tumor-suspicion prompt, then recursively scores the children of retained tiles; at each level, the top-$K$ tiles classified ``Yes'' are retained per parent. A tile is tumor-positive when more than $5\%$ of its area overlaps the pathologist-annotated tumor region. Conditional hit rate measures recall among positive tiles reachable from branches retained at the preceding level, whereas unconditional hit rate measures recall among all positive tiles at that magnification, so unreached branches count as misses. Both rates micro-average tile counts across the cohort. The protocol uses a retention schedule of $(6, 3, 2)$ at $(2.5\times, 10\times, 40\times)$ on the 190-WSI breast cohort. We additionally compare non-VLM baselines and a branching ($K$) sweep with a ground-truth oracle (Table~\ref{tab:ablation_t3b}).

\subsubsection{Evidence Integration}

\begin{problembox}
\noindent\textbf{Q.} Can a vision-language model integrate the per-magnification findings along a diagnostic route into a single diagnostic conclusion?
\end{problembox}

As a path-level probe of the reasoning term in Eq.~\eqref{eq:decomposition}, \emph{multi-scale diagnostic reasoning} presents the organ, the three pathologist-authored findings along a diagnostic route, and a four-option Diagnose, Triage, or Refine question. For \emph{Diagnose}, the correct option is the pathologist-composed route summary and distractors come from other paths of the same organ; \emph{Triage} maps the summary to a broader pathological category, whereas \emph{Refine} targets a fine-grained detail such as subtype, invasion, grade, or marker. The text-only input isolates integration from upstream perception and acquisition. Distractors are selected using the same \texttt{all-MiniLM-L6-v2}-based greedy low-similarity procedure as in T1/T2, with a cosine-similarity threshold of $0.6$. \emph{Triage} and \emph{Refine} stems are LM-generated with retry-and-validate guards. The released task contains $51{,}167$ four-option questions.

\section{Experimental Results and Analyses}
\label{sec:experiments}

\subsection{Experimental Setup}
\label{sec:setup}

We evaluate 20 model configurations: 19 VLMs stratified by training domain (general-purpose / pathology-specialised / general-medical) and one text-only reference model. MoE rows report total / active parameters, while undisclosed closed-source sizes use public analyst estimates. The general-purpose group comprises two closed-source VLMs, GPT-5.2~\cite{openai_update_2025} and Gemini-3-Flash~\cite{googledeepmindGemini3Flash2025}, together with the open-weight VLMs Qwen-3.5-Flash~\cite{qwenteamQwen35NativeMultimodal2026}, Kimi-K2.5~\cite{teamKimiK25Visual2026}, InternVL2.5-8B and InternVL2.5-26B~\cite{chenExpandingPerformanceBoundaries2025}, LLaMA-3.2-11B-Vision~\cite{meta2024llama32}, Qwen2.5-VL-7B~\cite{baiQwen25VLTechnicalReport2025}, Phi-3-Vision-4.2B~\cite{abdinPhi3TechnicalReport2024}, LLaVA-1.5-7B~\cite{Liu_2024_CVPR}, and LLaMA3-LLaVA-Next-8B~\cite{li2024llavanextstrong}. We additionally include the text-only DeepSeek-V4-Flash~\cite{deepseek-aiDeepSeekV4HighlyEfficient2026}. The pathology-specialised group consists of Patho-R1-7B~\cite{zhangPathoR1MultimodalReinforcement2026} and Quilt-LLaVA-7B~\cite{Seyfioglu_2024_CVPR}. The general-medical group comprises LLaVA-Med-1.5-7B~\cite{NEURIPS2023_5abcdf8e}, MedGemma-1.5-4B, MedGemma-4B, and MedGemma-27B~\cite{sellergrenMedGemmaTechnicalReport2026}, together with Lingshu-7B and Lingshu-32B~\cite{lasateamLingshuGeneralistFoundation2025}. The T4 sub-columns decompose the clinical reasoning chain: \emph{Triage} (broad pathology class), \emph{Diagnose} (integrated conclusion), and \emph{Refine} (subtype, grade, marker, or morphology).

Table~\ref{tab:results} reports the full T1/T2/T4 benchmarks, the fixed 50-slide T3A subset (150 queries per model), and the full 190-slide T3B cohort; Tables~\ref{tab:ablation}--\ref{tab:ablation_t3b} report their respective ablation subsets.

Table~\ref{tab:results} reports the full T1/T2/T4 benchmarks together with the Task~3 evaluation cohorts; Tables~\ref{tab:ablation}--\ref{tab:ablation_t3b} report their respective ablation subsets.

Model coverage varies with interface compatibility and computational cost. Mode~A evaluates GPT-5.2, Gemini-3-Flash, Qwen-3.5-Flash, and Patho-R1, although Patho-R1 returned no valid localization outputs. Kimi-K2.5 was evaluated only on T1 because of runtime, DeepSeek-V4-Flash only on T4 because it is text-only, and Phi-3-Vision and LLaMA-3.2-11B-Vision were omitted from T2 because they cannot accept four images. Other Mode~B omissions reflect runtime or uninformative outputs.

\begin{table*}[tp]
\centering
\caption{Benchmark-construction ablations for distractor encoder (T1/T2) and magnification inputs (T4). \textbf{Bold} marks the easiest encoder per model/task or the best single magnification or magnification pair.}
\label{tab:ablation}
\renewcommand{\arraystretch}{0.98}
\scriptsize
\setlength{\tabcolsep}{3.0pt}
\begin{tabular*}{\textwidth}{@{\extracolsep{\fill}} l c c c c c}
\toprule
\textbf{Model} & \textbf{Encoder--T1} & \textbf{Encoder--T2} & \textbf{Single magnification} & \textbf{Magnification pair} & \textbf{All three} \\
\cmidrule(lr){2-2}\cmidrule(lr){3-3}\cmidrule(lr){4-4}\cmidrule(lr){5-5}\cmidrule(lr){6-6}
& MiniLM/BiomedCLIP/CONCH & MiniLM/BiomedCLIP/CONCH & 2.5/10/40 & 2.5+10/10+40/2.5+40 & 2.5+10+40 \\
\midrule
\multicolumn{6}{@{}l}{\textit{General-purpose models}} \\
GPT-5.2 & 61.2/\textbf{66.8}/59.1 & 63.4/\textbf{64.1}/62.6 & 82.6/\textbf{85.5}/83.7 & 88.5/88.5/\textbf{88.7} & 90.5 \\
Gemini-3-Flash & \textbf{73.3}/70.8/70.5 & 68.0/70.5/\textbf{71.6} & 83.7/\textbf{87.7}/85.3 & 89.8/89.5/\textbf{89.9} & 91.6 \\
Qwen-3.5-Flash & \textbf{65.6}/60.3/63.3 & 62.5/\textbf{65.5}/64.9 & 85.7/\textbf{88.3}/85.9 & 90.3/90.6/\textbf{91.0} & 92.7 \\
InternVL2.5-26B & \textbf{56.0}/45.3/45.9 & 37.2/\textbf{38.3}/36.6 & 86.3/\textbf{90.9}/88.2 & 92.2/\textbf{92.8}/92.6 & 94.5 \\
InternVL2.5-8B & \textbf{58.3}/44.9/44.4 & 33.8/\textbf{37.2}/32.5 & 88.1/\textbf{90.9}/89.6 & 92.1/\textbf{92.3}/\textbf{92.3} & 93.2 \\
LLaMA3-LLaVA-Next-8B & \textbf{41.0}/31.1/35.9 & 24.8/\textbf{26.1}/24.6 & 77.5/\textbf{82.7}/81.5 & 86.1/\textbf{86.9}/85.3 & 88.5 \\
LLaVA-1.5-7B & \textbf{31.6}/26.8/28.8 & \textbf{25.7}/\textbf{25.7}/\textbf{25.7} & 37.7/37.2/\textbf{38.1} & 43.0/43.3/\textbf{44.3} & 43.3 \\
Qwen2.5-VL-7B & \textbf{52.0}/42.2/41.6 & \textbf{28.9}/27.9/28.7 & 78.1/\textbf{83.7}/79.2 & 85.3/\textbf{86.1}/84.3 & 87.3 \\
\midrule
\multicolumn{6}{@{}l}{\textit{Pathology-specialised VLMs}} \\
Patho-R1-7B & \textbf{43.7}/31.6/38.2 & 24.6/\textbf{25.9}/24.8 & 38.5/45.0/\textbf{49.5} & 44.1/\textbf{47.8}/44.5 & 51.8 \\
Quilt-LLaVA-7B & 10.5/9.5/\textbf{12.4} & 8.3/\textbf{8.9}/8.7 & 60.3/\textbf{60.9}/59.8 & \textbf{67.5}/65.1/65.6 & 67.1 \\
\midrule
\multicolumn{6}{@{}l}{\textit{General-medical VLMs}} \\
MedGemma-27B & \textbf{47.4}/38.7/42.1 & 29.1/30.5/\textbf{30.6} & 84.0/\textbf{89.5}/84.7 & \textbf{90.9}/90.5/\textbf{90.9} & 93.1 \\
Lingshu-7B & \textbf{69.8}/60.7/62.6 & 58.2/59.5/\textbf{60.2} & 87.7/\textbf{91.5}/89.3 & 92.1/\textbf{92.9}/92.1 & 93.8 \\
LLaVA-Med-1.5-7B & \textbf{27.7}/27.6/23.2 & \textbf{26.1}/\textbf{26.1}/26.0 & 59.2/\textbf{66.9}/64.0 & 62.5/\textbf{66.9}/63.7 & 65.0 \\
MedGemma-1.5-4B & \textbf{53.3}/41.8/46.7 & 36.8/\textbf{38.8}/\textbf{38.8} & 66.4/\textbf{71.9}/70.3 & \textbf{73.3}/73.0/71.3 & 75.4 \\
MedGemma-4B & \textbf{55.9}/39.6/40.4 & 35.4/\textbf{36.1}/35.3 & 82.9/\textbf{86.7}/86.3 & 88.6/\textbf{89.0}/87.2 & 90.2 \\
\bottomrule
\end{tabular*}
\vspace{2pt}
\parbox{\textwidth}{\tiny\textit{Evaluation subsets:} T1 and T2 use 1{,}071 questions each; T4 uses 1{,}500 questions.}
\end{table*}

\begin{table}[tp]
\centering
\caption{Task-3 Mode~A navigation ablation. Exploration-budget sweep: mean IoU per magnification under a \emph{Tight}, the default \emph{Base}, and a \emph{Loose} ROI budget.}
\label{tab:ablation_t3a}
\renewcommand{\arraystretch}{1.0}
\scriptsize
\setlength{\tabcolsep}{5pt}
\begin{tabular*}{\columnwidth}{@{\extracolsep{\fill}} l ccc@{}}
\toprule
\textbf{Budget} & 2.5$\times$ & 10$\times$ & 40$\times$ \\
\midrule
\multicolumn{4}{@{}l}{\textit{GPT-5.2}} \\
~~Tight (5,3,2) & 0.002 & 0.007 & 0.000 \\
~~\textbf{Base (10,7,5)} & \textbf{0.086} & \textbf{0.059} & \textbf{0.066} \\
~~Loose (20,15,10) & 0.006 & 0.033 & 0.003 \\
\midrule
\multicolumn{4}{@{}l}{\textit{Gemini-3-Flash}} \\
~~Tight (5,3,2) & 0.007 & 0.006 & 0.000 \\
~~\textbf{Base (10,7,5)} & \textbf{0.041} & \textbf{0.089} & \textbf{0.068} \\
~~Loose (20,15,10) & 0.043 & 0.023 & 0.012 \\
\midrule
\multicolumn{4}{@{}l}{\textit{Qwen-3.5-Flash}} \\
~~Tight (5,3,2) & 0.005 & 0.022 & 0.010 \\
~~\textbf{Base (10,7,5)} & \textbf{0.051} & \textbf{0.079} & \textbf{0.083} \\
~~Loose (20,15,10) & 0.017 & 0.013 & 0.011 \\
\bottomrule
\end{tabular*}
\vspace{2pt}
\parbox{\columnwidth}{\tiny\textit{Evaluation subset:} 50 slides with three queries per slide (150 queries per model per budget).}
\end{table}

\begin{table*}[tp]
\centering
\caption{Task-3 Mode~B branching-schedule ablation. Directly measured unconditional hit rate (\%) is reported at 2.5$\times$, 10$\times$, and 40$\times$; \textbf{bold} marks the default $(K_{2.5},K_{10},K_{40})=(6,3,2)$.}
\label{tab:ablation_t3b}
\renewcommand{\arraystretch}{1.0}
\scriptsize
\setlength{\tabcolsep}{2.2pt}
\begin{tabular*}{\textwidth}{@{\extracolsep{\fill}}l ccc ccc ccc ccc ccc@{}}
\toprule
 & \multicolumn{3}{c}{Random} & \multicolumn{3}{c}{Patho-R1} & \multicolumn{3}{c}{Qwen} & \multicolumn{3}{c}{Gemini} & \multicolumn{3}{c}{Oracle} \\
\cmidrule(lr){2-4}\cmidrule(lr){5-7}\cmidrule(lr){8-10}\cmidrule(lr){11-13}\cmidrule(lr){14-16}
$(K_{2.5},K_{10},K_{40})$ & \tiny2.5 & \tiny10 & \tiny40 & \tiny2.5 & \tiny10 & \tiny40 & \tiny2.5 & \tiny10 & \tiny40 & \tiny2.5 & \tiny10 & \tiny40 & \tiny2.5 & \tiny10 & \tiny40 \\
\midrule
(1,1,1) & 6.58 & 0.72 & 0.02 & 4.72 & 0.54 & 0.03 & 11.40 & 1.43 & 0.08 & 11.40 & 1.38 & 0.08 & 17.98 & 2.20 & 0.14 \\
(3,3,3) & 18.42 & 4.74 & 0.64 & 10.37 & 1.81 & 0.23 & 28.95 & 10.08 & 1.65 & 31.58 & 11.13 & 1.82 & 42.98 & 15.87 & 2.87 \\
\textbf{(6,3,2)} & 30.70 & 9.15 & 0.87 & 7.92 & 2.92 & 0.30 & 51.75 & 18.46 & 2.06 & 54.82 & 19.89 & 2.24 & 67.11 & 24.68 & 3.01 \\
(6,5,3) & 33.77 & 12.62 & 1.84 & 12.40 & 3.76 & 0.54 & 51.75 & 27.93 & 4.62 & 70.37 & 38.99 & 5.92 & 67.54 & 37.47 & 6.78 \\
(10,5,3) & 47.37 & 18.18 & 2.52 & 9.84 & 3.19 & 0.43 & 69.74 & 37.63 & 6.25 & --- & --- & --- & 84.21 & 45.84 & 8.26 \\
\bottomrule
\end{tabular*}
\vspace{2pt}
\parbox{\textwidth}{\tiny\textit{Evaluation subset:} 45 slides. Gemini was not evaluated under $(K_{2.5},K_{10},K_{40})=(10,5,3)$ because of excessive API waiting time.}
\end{table*}

\subsection{Patch-Level Recognition (Tasks 1 and 2)}
\label{sec:patch}

Tasks~1 and 2 share samples and distractor pools (Table~\ref{tab:results}). Gemini-3-Flash leads both (T1 63.5\%, T2 67.7\%), and the strongest five T1 models span 52.2--63.5\%. The pathology-specialised checkpoints fall below the 25\% random baseline: Patho-R1 reaches only 20--22\%, while Quilt-LLaVA falls to about 7\%. The general-medical LLaVA-Med similarly reaches only 19--20\%. The distractor-encoder columns in Table~\ref{tab:ablation} show task-dependent score shifts but stable model ordering (Spearman $\rho=0.95$--$0.99$), indicating that the ranking is not MiniLM-specific.

\subsection{Diagnostic Region Localization (Task 3)}
\label{sec:task3_results}

Both Task~3 regimes remain challenging, but the two interfaces assign different roles to the VLM (Table~\ref{tab:results}). In Mode~A the VLM directly controls region selection through tool calls, and no evaluated VLM exceeds IoU~$0.09$: a parameter-free parent-box-center heuristic (IoU $0.25$--$0.28$ at 10$\times$/40$\times$) beats every VLM by 3--4$\times$, Patho-R1 fails to emit valid bounding-box calls (IoU~$\equiv 0$), and \emph{enlarging} the exploration budget degrades IoU further---the sweep is an inverted~U (Table~\ref{tab:ablation_t3a}). In Mode~B the VLM is restricted to per-tile scoring inside a fixed hierarchical top-$K$ pipeline; on the controlled $K$-sweep subset, Gemini and Qwen reach 54.8\% and 51.8\% unconditional hit rate at 2.5$\times$, versus 30.7\% for random sampling. Thus, current VLMs score evidence more reliably than they plan navigation.

Constrained scoring still suffers severe end-to-end attrition: Gemini's directly measured unconditional hit rate falls $52.2\%\!\to\!18.5\%\!\to\!2.02\%$ from 2.5$\times$ to 40$\times$, and Patho-R1-7B remains below random under the default schedule. The controlled sweep now exposes the same mid-scale loss directly: under $(6,3,2)$, Gemini falls $54.82\%\!\to\!19.89\%\!\to\!2.24\%$ and Qwen falls $51.75\%\!\to\!18.46\%\!\to\!2.06\%$. Branching budget is a real but incomplete remedy (Table~\ref{tab:ablation_t3b}): expanding Qwen from $(6,3,2)$ to $(10,5,3)$ raises 40$\times$ coverage from 2.06\% to 6.25\%, while the ground-truth oracle rises from 3.01\% to only 8.26\%. The bottleneck therefore couples early pruning with scorer quality; retaining more tiles buys recall with compute but still does not yield the precise localization a clinical workflow needs.

\begin{figure*}[t]
\centering
\includegraphics[width=0.96\textwidth]{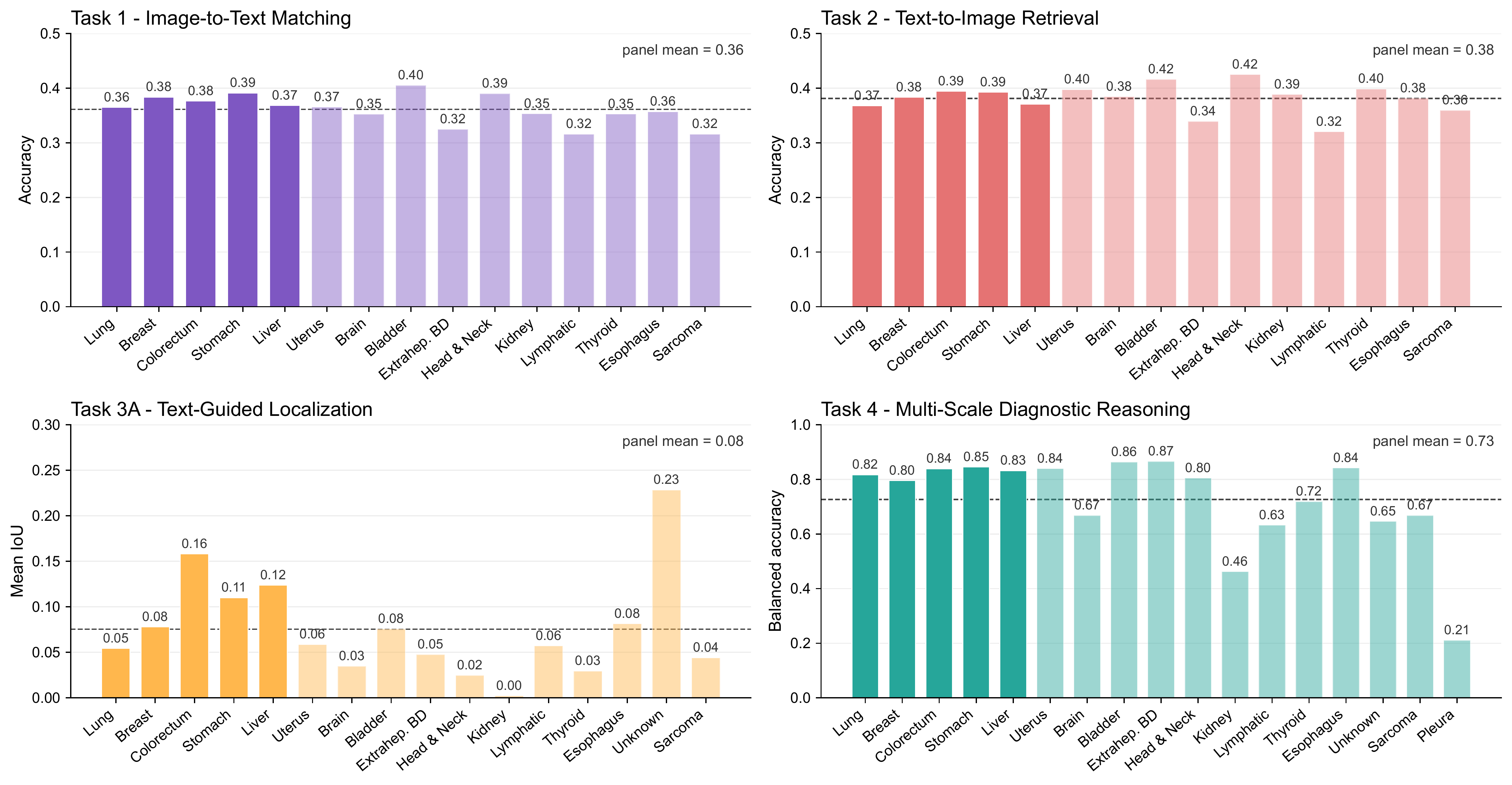}
\caption{Per-organ performance across the four \benchname{} tasks, averaged over the task-specific evaluated models and baselines. Mode~B is omitted because its breast-only cohort has no organ axis. Saturated and faded bars denote high-coverage and rare-tail organs, respectively; dashed lines mark unweighted organ means.}
\label{fig:organ_breakdown}
\end{figure*}

\subsection{Multi-Scale Diagnostic Reasoning (Task 4)}
\label{sec:task4_results}

T4 is easy once observations are supplied (Table~\ref{tab:results}): twelve models exceed 87\%, with the top three open-weight $\leq$26\,B checkpoints (InternVL2.5-8B 93.4\%, Lingshu-7B 93.4\%, InternVL2.5-26B 93.2\%) beating every closed-source VLM. Among the strongest models, most remaining errors occur on \emph{Diagnose}: broad triage is nearly saturated, but deriving the integrated conclusion is less reliable. Multi-scale evidence generally helps. For 15 of the 16 models in the magnification ablation, the full three-scale input exceeds the best single magnification by 2.3--6.3\,pt. The 10$\times$ view is the strongest single scale for 14 models, whereas Patho-R1 and LLaVA-1.5 perform best at 40$\times$. LLaVA-Med is the sole exception to the multi-scale gain, scoring 66.9\% at 10$\times$ versus 65.0\% with all three scales. Pathology-specialised models span 56--70\%, while LLaVA-1.5 remains a 40.6\% prompt-format outlier.

\subsection{Closed-Source Inference Efficiency}
\label{sec:closed_source_efficiency}

We further evaluate the inference efficiency of GPT-5.2 and Gemini-3-Flash on shared fixed subsets comprising 20 questions for Tasks~1, 2, and 4, five slides at each magnification for Task~3 Mode~A, and five slides for Mode~B. Mean end-to-end latency for GPT-5.2/Gemini-3-Flash was 20.1/22.3\,s per question for Task~1, 68.0/94.5\,s for Task~2, and 1.7/2.7\,s for Task~4. Across models and magnifications, Mode~A required 41.0--221.6\,s per slide at a given magnification, whereas Mode~B required 20.7/46.8\,min per slide, showing that iterative evidence acquisition dominates inference time. Cost followed the same trend, increasing from \$0.00025--\$0.00284 per MCQ question to \$0.195--\$0.224 per Mode~B slide.

\subsection{Per-Organ Robustness}
\label{sec:organ_breakdown}

Across the five high-coverage organs (Lung, Breast, Colorectum, Stomach, and Liver), per-organ means remain tightly grouped within each MCQ task, with max–min ranges of 2.6\,pt on both T1 and T2 and 5.0\,pt on T4 (Fig.~\ref{fig:organ_breakdown}). Colorectum is the only clear high-coverage outlier on T3A (IoU\,$0.16$ vs.~the $0.08$ panel mean). Larger deviations occur mainly among rare-tail organs with very small evaluation sets, including Sarcoma ($n{=}4$ on T1/T2) and Pleura ($n{=}2$ on T4). Thus, the task-axis bottlenecks are consistent across the high-coverage cohort but remain less certain for sparsely sampled organs.

\subsection{Cross-Task Findings}
\label{sec:cross_task}

\begin{figure*}[t]
\centering
\begin{subfigure}[t]{0.57\textwidth}
\vspace{0pt}
\centering
\makebox[\linewidth][r]{%
\includegraphics[width=1.143\linewidth,keepaspectratio]{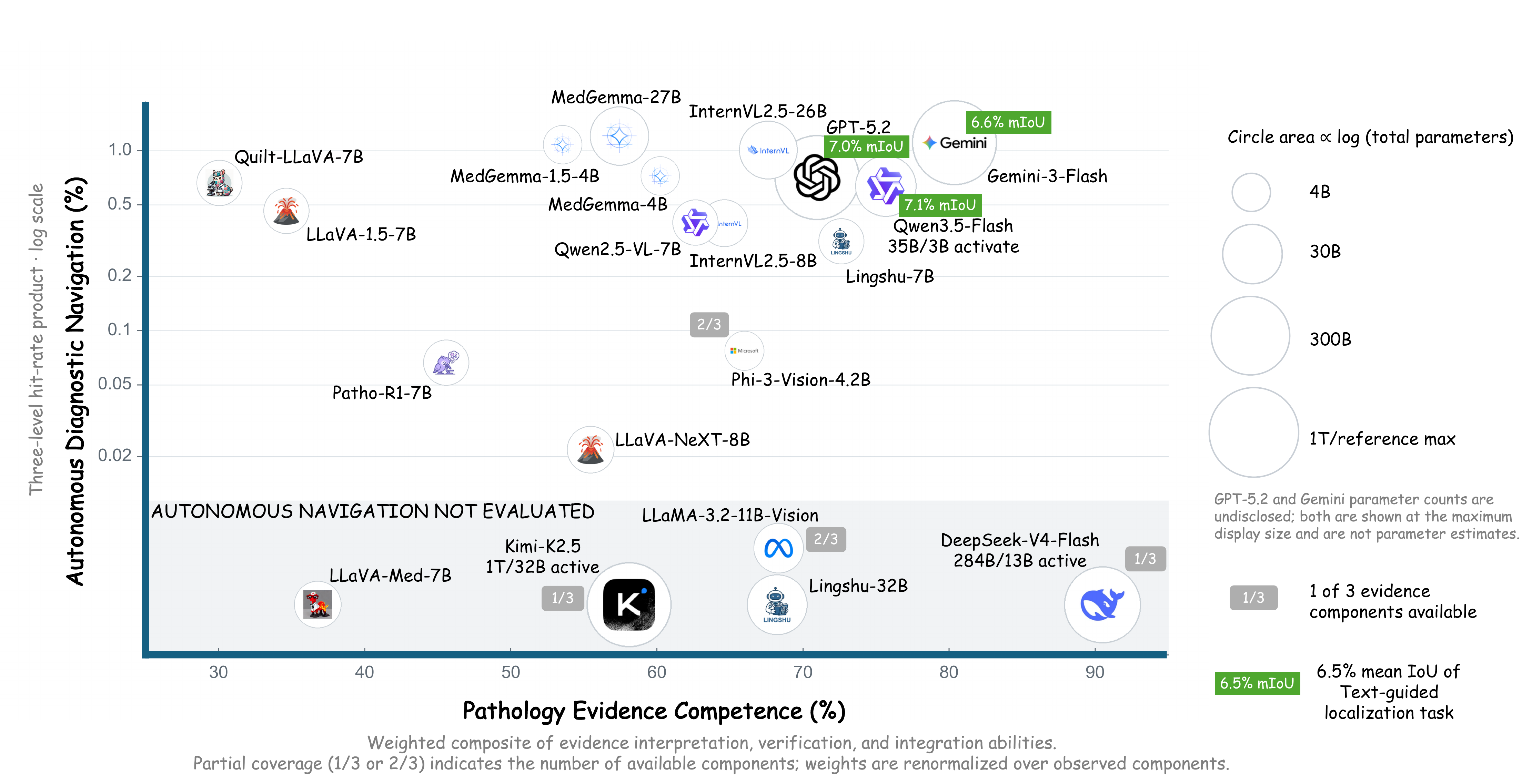}}
\caption{Cross-task capability landscape. The horizontal axis combines Tasks~1, 2, and 4 (25\%/25\%/50\%); the vertical axis is the product of the three conditional Task~3 Mode~B hit rates (log scale). Marker area encodes total parameters; the shaded band denotes unavailable Mode~B results.}
\label{fig:cross_task}
\end{subfigure}\hfill
\begin{subfigure}[t]{0.41\textwidth}
\vspace{0pt}
\centering
\makebox[\linewidth][l]{%
\includegraphics[width=1.11\linewidth,keepaspectratio]{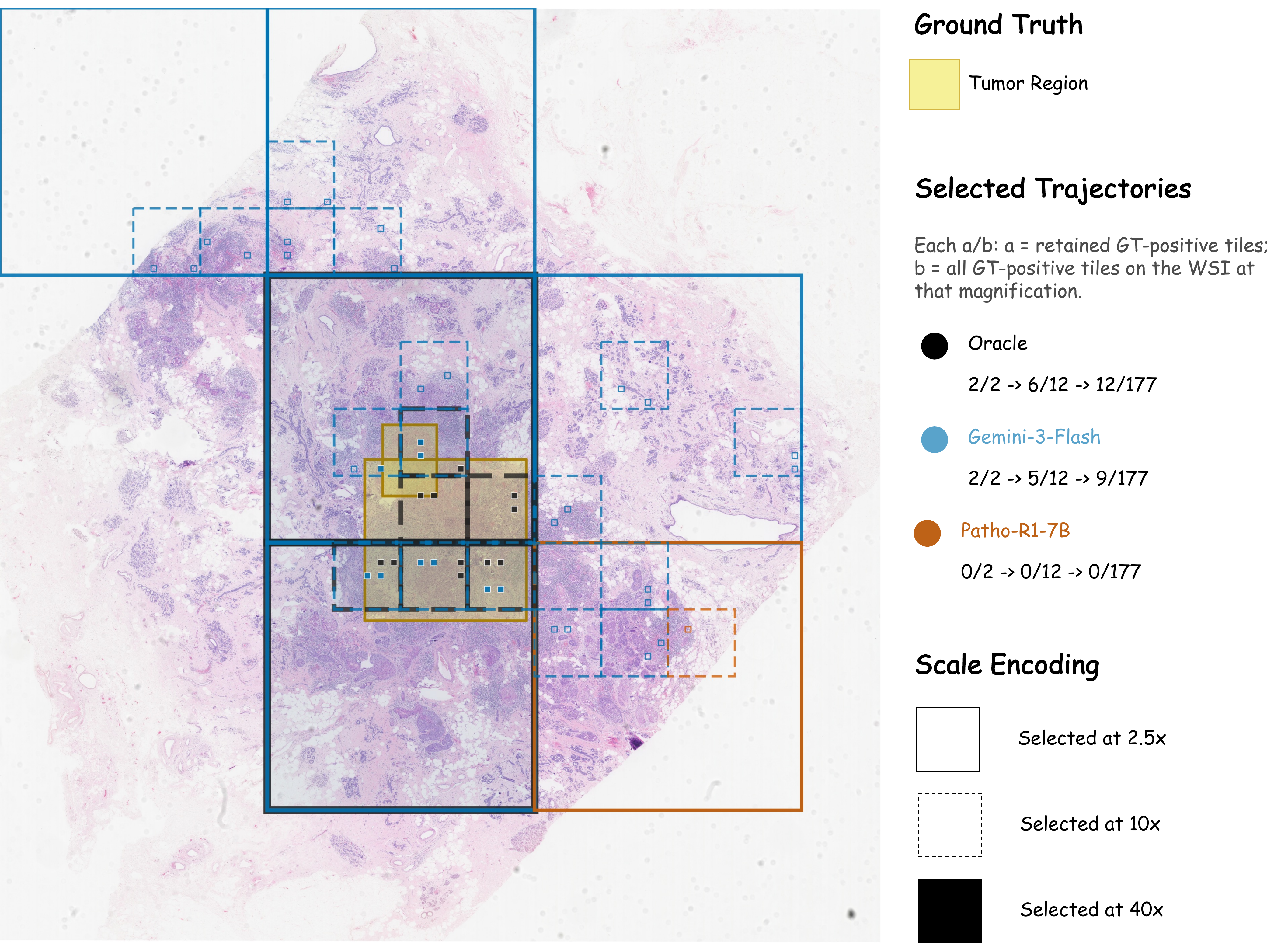}}
\caption{Task~3 Mode~B navigation on one WSI. Yellow marks pathologist-annotated tumor ground truth; black, blue, and orange boxes show the multi-scale regions retained by the GT-scored oracle, Gemini-3-Flash, and Patho-R1-7B.}
\label{fig:task3b_case_study}
\end{subfigure}
\caption{Cross-task performance summary and representative Task~3 Mode~B navigation.}
\vspace{-6pt}
\end{figure*}

We distil the per-task results into four findings; Fig.~\ref{fig:cross_task} summarises the per-model profiles.

\textit{Finding 1 -- Once multi-scale evidence is supplied, integration is no longer the bottleneck.} The expert reference sharpens this stage-wise asymmetry. On the reported expert subsets, pathologists reached 93.6\% and 92.3\% on T1 and T2, whereas the best VLM results were 63.5\% and 67.7\%, respectively. In contrast, the gap narrowed to 96.6\% versus 93.4\% on T4. Current VLMs therefore approach expert-reference accuracy only after multi-scale evidence has been selected and verbalised, while visual evidence interpretation and verification remain substantially weaker. This distinction also shows why performance on supplied-evidence reasoning tasks can overstate end-to-end diagnostic readiness.

\textit{Finding 2 -- Patch-level recognition generally improves with magnification.} In Table~\ref{tab:results}, 17 of 19 evaluated models on T1 and 11 of 16 on T2 achieve their highest accuracy at 40$\times$. This pattern is also evident in the leading model: Gemini-3-Flash improves from 57.3\% to 64.6\% on T1 and from 62.8\% to 69.8\% on T2 between 2.5$\times$ and 40$\times$. Thus, both image-to-text interpretation and text-to-image verification generally benefit from fine-scale morphological detail.

\textit{Finding 3 -- Evidence acquisition remains the primary bottleneck in hierarchical search.} Mode~A shows that target guidance alone cannot produce reliable spatial plans, while Mode~B shows that restricting the model to tile scoring improves early selection but not fine localization; retaining more candidates recovers recall without overcoming errors made early in an irrevocable search, so interface constraints help without solving acquisition.

\textit{Finding 4 -- Pathology-specific training alone does not guarantee better task performance.} Among 7\,B models, the general-medical Lingshu-7B clearly outperformed Patho-R1 and Quilt-LLaVA on T1, T2, and T4, while LLaVA-Med performed at a similarly low level. Thus, medical or pathology-specific training alone does not ensure strong performance on these tasks. Localization also requires models to ground spatial information, use tools effectively, and rank candidate regions reliably. Neither a larger model nor pathology-focused training alone solves evidence acquisition.

\section{Discussion}
\label{sec:discussion}

\subsection{Interpreting the Cross-Task Gap}
\label{sec:interpretation}

Taken together, the results suggest that end-to-end WSI competence is limited not by a single missing capability, but by the composition of partially competent stages. Task~4 evaluates reasoning after multi-scale evidence has already been selected and verbalised, whereas Tasks~1 and 2 still require visual--language binding, and Task~3 embeds perception within a sequential evidence-acquisition policy. Strong performance on supplied-evidence reasoning can therefore coexist with weak visual interpretation and localization, and should not be taken as evidence of readiness for autonomous pathology workflows.

The two Task~3 modes further localise where this composition breaks down. Mode~A requires a model to plan spatially, transform coordinates, invoke tools, and decide when to stop; additional exploration can amplify spatial drift rather than improve localization. Mode~B removes most planning demands, but repeated top-$K$ selection converts imperfect tile ranking into cumulative coverage loss. These distinct failure modes argue for evaluating region scoring, spatial planning, tool use, and stopping separately before recombining them, and for using trajectory-level supervision to identify when and why a search departs from a diagnostically useful path.

The T2-over-T1 asymmetry among the leading models is consistent with explicit textual hypotheses narrowing the visual search space. Although this comparison does not by itself establish the superiority of a particular search policy, it motivates hypothesis-driven evidence acquisition: a model could formulate a differential diagnosis, retrieve regions relevant to explicit morphological hypotheses, and revise its search when the retrieved evidence conflicts with them. Such a workflow would create interpretable checkpoints between evidence proposal, verification, and diagnostic revision.

\subsection{Design and Deployment Implications}\label{sec}

The results indicate that evidence acquisition must be trained directly rather than assumed to emerge from pathology knowledge or diagnostic instruction tuning. Useful curricula should combine hard visual--text negatives, multi-image and cross-region verification, coordinate grounding, structured tool use, and supervised navigation trajectories. Corresponding evaluations should report stage-specific errors in addition to terminal localization performance, so that improvements in recognition are not conflated with improvements in search.

WSI search policies should also be reversible. Under irrevocable hierarchical pruning, even a locally plausible error can eliminate all diagnostically relevant descendants. Beam search, sibling reconsideration, and backtracking provide mechanisms for recovery, but they depend on ranking signals that distinguish strong evidence from ambiguous candidates. Reliable confidence ranking is therefore a structural requirement for adaptive search, rather than merely a calibration refinement applied after navigation.

Finally, evidence acquisition must be evaluated as a deployment problem as well as a modelling problem. The measured latency and cost of iterative WSI navigation, together with repeated transmission of image tiles and clinical context to external services, raise concerns about turnaround time, scalability, and privacy. On-premises and hybrid configurations should therefore be compared under realistic hardware constraints using end-to-end latency, throughput, cost, and data-governance criteria.

\subsection{Limitations}
\label{sec:limitations}

The autonomous-navigation protocol currently covers only breast-cancer WSIs, so generalisation across organs and tissue architectures remains untested. The restricted Mode~A model panel also limits generalisability beyond VLMs capable of reliable tool-mediated localization. Both localization modes use fixed budgets or branching schedules, and adaptive policies may change absolute performance. Although findings and diagnoses are pathologist-authored, distractors rely on MiniLM similarity; rank stability across the encoder ablation reduces, but does not eliminate, this concern. The 93\% T4 result is also bounded by the four-option format and should not be interpreted as open-ended diagnostic competence. Survival prediction, treatment response, and molecular subtyping remain outside the diagnostic-tree decomposition.

\section{Conclusion}
\label{sec:conclusion}

In this work, we introduced \benchname{}, a diagnostic-tree benchmark for evaluating evidence acquisition and reasoning in pathology VLMs. Across 20 model configurations, leading systems exceed 93\% accuracy when integrating supplied multi-scale observations, yet text-guided localization remains below 0.09 mean IoU and autonomous tumor coverage falls from 52.2\% at 2.5$\times$ to 2.02\% at 40$\times$. This asymmetry redirects future research from stronger reasoning over curated evidence toward reliable evidence acquisition. Progress requires spatially grounded training for coordinate reasoning and tool use; confidence-calibrated adaptive search with backtracking to recover from pruning errors; and hybrid verification with segmentation or task-specific detectors. These mechanisms should be tested across organs and in open-ended diagnostic settings, alongside privacy-preserving deployment with acceptable end-to-end latency and cost, before VLM-based WSI systems are considered clinically ready.

{
\renewcommand{\footnotesize}{\fontsize{7.9}{8.7}\selectfont}
\bibliographystyle{IEEEtran}
\bibliography{references}
}

\end{document}